\documentclass[preprint,12pt,authoryear]{elsarticle}

\usepackage{setspace}
\usepackage{amssymb}
\usepackage{booktabs}
\usepackage{url}
\usepackage{graphicx}
\usepackage{algorithm}
\usepackage{algorithmic} 
\usepackage{subcaption}
\usepackage[load-configurations=version-1]{siunitx} 

\journal{Neurocomputing}

\begin{document} 
\raggedright

\begin{frontmatter}

\title{Multi-level conformal clustering: \linebreak
	 A distribution-free technique for clustering and anomaly detection}

\author[add1,addc*]{Ilia Nouretdinov}

\author[add2]{James Gammerman}

\author[add3]{Matteo Fontana}

\author[add2]{Daljit Rehal}

\author[]{Declarations of interest: none}

\address[add1]{Computer Learning Research Center, Royal Holloway University of London, UK}
\address[add2]{Centrica plc, UK}
\address[add3]{Department of Management, Economics and Industrial Engineering, Politecnico di Milano, Italy}
\address[addc*]{Corresponding author, e-mail: I.R.Nouretdinov@rhul.ac.uk}

\begin{abstract}

In this work we present a clustering technique called \textit{multi-level conformal clustering (MLCC)}. The technique is hierarchical in nature because it can be performed at multiple significance levels which yields greater insight into the data than performing it at just one level. We describe the theoretical underpinnings of MLCC, compare and contrast it with the hierarchical clustering algorithm, and then apply it to real world datasets to assess its performance. There are several advantages to using MLCC over more classical clustering techniques: Once a significance level has been set, MLCC is able to automatically select the number of clusters. Furthermore, thanks to the conformal prediction framework the resulting clustering model has a clear statistical meaning without any assumptions about the distribution of the data. This statistical robustness also allows us to perform clustering and anomaly detection simultaneously. Moreover, due to the flexibility of the conformal prediction framework, our algorithm can be used on top of many other machine learning algorithms.
\end{abstract}


\end{frontmatter}

\section{Introduction}

The aim of this paper is to present an extension to conformal clustering (CC), a clustering and anomaly detection technique recently developed in ~\cite{Cherubin2015}. The original paper did not explain or examine the technique in any detail, which is the first objective of this paper. We then show how it can be extended to run at a range of significance levels to allow for fine tuning of the clustering. \linebreak

In the traditional conformal prediction (CP) setting as described in \cite{Gammerman2005}, we are presented with a training set of observations (also called examples), each consisting of an object and its label. We are then presented with a new test object and asked to predict its label. We train a conformal predictor to produce a prediction set (often with multiple labels, or even none) depending on the level of confidence we want to have in our predictions. \linebreak
 
CC consists of applying the CP framework to the unsupervised learning task of clustering. However, instead of predicting labels for new test objects, we evaluate the whole feature space (even those parts of the space where we do not observe data) and identify which areas of it `conform' to the distribution of the observed data. The advantage of this conformal approach is that it is distribution-free: that is, the only assumption applied to the distribution of the data is that it satisfies the \textit{i.i.d} condition. This make this approach different from clustering models involving mixtures of Gaussians or other assumptions as in \cite{Liu2008}.\linebreak

In CC no labels are involved at the clustering stage - we only deal with objects. This allows us to identify a so-called `region of conformity' (analogous to the prediction set in more traditional CP) in the feature space: data points which lie within this region of the feature space are grouped into different clusters, while those that lie outside it are classified as anomalies.  \linebreak

A key property of CC is its dependence on a pre-selected significance level $\varepsilon$. This parameter lets us tune the size of the clusters by changing their boundaries. As $\varepsilon$ increases, the more `abnormal' data points in a cluster are re-classified as anomalies, which reduces the size of the cluster and may cause it to split into two or more new clusters. In other words, we can use $\varepsilon$ to regulate both the level (depth) of clustering and the extent of anomaly detection.  \linebreak

In this paper on \textit{multi-level conformal clustering (MLCC)} we investigate that relationship, and in doing so introduce a new form of dendrogram analagous to the one produced in hierarchical clustering (HC). However, the dendrograms differ because MLCC does not force the merging of clusters when there is no actual clustering tendency. This is in contrast to agglomerative HC where clusters are forcibly merged together until the whole dataset is one large cluster. In MLCC if there is no reason for two clusters to be merged (because they are dissimilar) then they are not, and both (may) eventually disappear as the significance level gets higher. In this situation, data points which were previously located in those clusters are re-classified as anomalies. We are thus presenting a technique in which the $y$-axis of the dendrogram - that is, the level of clustering - has a statistical meaning, unlike in HC. A further advantage of the MLCC dendrogram is that the x-axis is easier to interpret than in HC: the more similar two observations are, the closer their proximity on the $x$-axis. \linebreak

In terms of related work, one approach for aggregating over many clustering levels was introduced in~\cite{Nugent2008}. 
However, this is different to our approach because the level of clustering was related to underlying density estimates, not to a significance level. 
Another study of density-based clustering for multi-dimensional data was also made in \cite{Rinaldo2010}, but not within a conformal framework, and not focusing on varying the significance level.
Clustering of multimodal distributions as a possible application of conformal prediction regions for was firstly mentioned in~\cite{lei2011}.
\linebreak

A previous study closer to ours which also uses a hierarchical tree and fixed significance levels (producing a so-called `conformal tree') was presented in~\cite{Lei2014}. However, this was only for the specific case of visualization and exploration of functional data (i.e. data embedded in a functional space). The objective was to develop exploratory "bands" for functions via the selection of specific nonconformity measures. We build on that study by extending its line of thought in three directions: firstly, we develop the ideas behind conformal trees into a clustering method comparable to hierarchical clustering, not merely an exploratory tool. Secondly, we develop our theoretical argument not only for the very specific case of functional data, but also for a standard multivariate setting.  Thirdly, by using a very general distance measure our method is able to produce clusters not only as bands, but as general shapes.\linebreak

This paper is structured as follows. We begin with a description and explanation of the MLCC methodology and how it relates to conformal prediction. Then we apply the methodology to both low and high-dimensional datasets before evaluating the technique and comparing it to HC. \linebreak

\section{Background Theory and Algorithm}

The theoretical background of CC and MLCC is based on a Conformal Prediction (CP) framework. In this section we provide a brief review of this framework.

\subsection{Conformal Prediction}\label{CC-intuition}

Conformal prediction \citep{Gammerman2005} is a framework for making reliable predictions without the use of complex probabilistic models. It involves making predictions as a set of arbitrary size rather than as single point values, and has theoretically proven guarantees of validity, with the only assumption being that the data are \textit{i.i.d}. \linebreak

Conformal prediction is applied within a machine learning setting. There is a sequence of objects $z_1,\dots,z_l$ and the task is to provide a prediction for a new example $z_{l+1}$. All examples belong to an object space $Z$. \linebreak

The CP framework can be applied to tasks from both supervised and unsupervised learning. In supervised learning, each $z_i\in Z$ is a pair $(x_i,y_i)$ consisting of a feature vector $x_i$ and a label $y_i$, so $Z$ is the product of a vector space $X$ and a label set $Y$.
In unsupervised learning (which includes tasks such as clustering and anomaly detection), $z_i$ typically just refers to the feature vector $x_i\in X$. 
In the next section on CC, when we refer to an example $z_i$ we mean only its feature vector.
\linebreak 

Conformal predictors can operate on top of another ML algorithm (known as the `underlying' algorithm). This can be any classification or regression algorithm - the only requirement is that we can extract a `score' from it, from which we can develop a \textit{nonconformity measure (NCM)}. The NCM evaluates numerically how different a new example (i.e. a feature vector - label relationship in the supervised learning setting, or just a feature vector in the case of unsupervised learning) is from a set of previous examples - that is, how non-conforming it is. The NCM can be thought of as a measure of how strange the new example is given what came before, such that we might exclude a label from our prediction set if it produces new examples with a high NCM.  \linebreak	

Formally, a NCM is a function $A : Z^{(*)} \times Z \rightarrow \mathbb{R} $ where 
$Z$ is the set  of all possible examples and 
$Z^{(*)}$ is the set of all bags of examples of $Z$. 
$A$ tells us how different an example $z_i\in Z$ is from a bag (multi-set) $Z^{(*)}$ by assigning it a score.
\linebreak 

We can give examples of possible NCMs by considering a simple example of an underlying algorithm such as {\it $k$-nearest neighbors}.
In the supervised setting, the strangeness of an example with respect to a bag may be the proportion of its $k$ nearest neighbors which have a different label to the example. Intuitively, if an example with one label is surrounded by examples with a different label then we can consider it to be strange or anomalous.

In the unsupervised setting, the NCM can be simplified to just the average distance between the examples and its nearest $k$ neighbors. This means that an example is strange/anomalous if
it is located in a low density area where the average distance to its neighbors is high. In this work we consider only the unsupervised setting.\linebreak 

More formally, consider now a new example $z_{l+1}$.
Our null hypothesis is that $z_{l+1}$ was drawn \textit{i.i.d} from the same distribution as the observed examples $z_1,\dots,z_l$.
Adding this new example $z_{l+1}$ to our dataset gives us an extended bag
$\{z_1,...,z_l,z_{l+1}\}$. \linebreak 

Now we can use a NCM to compute 
\begin{equation}
\alpha_i=A(\{z_1,...,z_{i-1}, z_{i+1},...,z_{l+1}\},z_i)
\end{equation}
for each example $i = 1,...,l+1$. 
Note that it is applied to each of the examples, not just for the new one. 
\linebreak

On their own the scores produced by the NCM for each example are not particularly informative. But if we compare $\alpha$ for example $z_{l+1}$ against that of all other examples then we can quantify how unusual it is by using the notion of a $p$-value:
\begin{equation}
p(z_{l+1}) = 
\frac{ \#\{i=1,...,l+1 : \alpha_i \geq \alpha_{l+1}\} } {l+1}
\end{equation}
This is $p$-value tests the null hypothesis stated above. Usually a threshold or \textit{significance level} $\varepsilon$ is selected, and if the $p$-value of example $z_{l+1}$ is higher than this threshold, then it is included into the prediction set.  \linebreak

\subsection{Conformal clustering} 

CC is an unsupervised technique in which all data objects are used for the training phase. We refer to this data as the \textit{observed dataset}. The objects are typically represented as \textit{feature vectors} and the space
of possible feature vectors can be called the \textit{virtual feature space}, which is usually a vector space. 
By `virtual' here we mean that some combinations of feature values are not observed in the data objects, but they are still covered by the feature space. \linebreak

We train a conformal predictor on the observed data, and then loop through every point of the virtual feature space (practically represented by a grid with a certain resolution) using each point as a quasi-test object: the NCM lets us calculate a $p$-value for each point of the grid that measures how similar it is to the observed data. \linebreak

The set of points which have a $p$-value above a given threshold (significance level) $\varepsilon$ is called the \textit{region of conformity}. That is,
a region of conformity in the feature space is the locus of points that conform well to the distribution of the observed data, in the sense that we do not reject the hypothesis of being generated by the same i.i.d. distribution.
This region will often consist of several clusters, or to use a term from topology and graph theory - \textit{connected components}. Throughout this report we will refer to connected components of the region of conformity as \textit{virtual clusters} because they are located in the virtual feature space.  \linebreak
 
Finally the actual observations are projected onto the feature space and classified according to whether they lie within these virtual clusters, forming corresponding \textit{data clusters}, while examples left outside the prediction set are classified as anomalies. A key advantage of CC is that we have the ability to toggle the proportion of data points that are classified as anomalies. \linebreak

To summarise, the steps to implement this algorithm are as follows:

\begin{enumerate}
	\item Train a conformal predictor to produce a region of connected components, or region of conformity, on a grid covering the whole feature space.
	
	\item Formally identify virtual clusters in the region of conformity.
	
	\item Project the observed data objects onto the grid. Identify data clusters and anomalies.
	
\end{enumerate}

Traditional CP is concerned with the notion of \textit{validity}, which provides guarantees against erroneous label prediction. In the context of CC we suggest that validity means the guarantee against a `false alarm', if we consider each classification of a data point as an anomaly as being a kind of alarm. This means that for a new data point  generated from the same distribution as the observed data, the chance of it being wrongly classified as an anomaly is limited by the value of the significance level. So for example, at a significance level of 0.05 we might say that with 95\% confidence a new data point will not be incorrectly classified as an anomaly, but instead be correctly placed inside the region of conformity. \linebreak

Note that such false alarms can cause clustering mistakes in two opposing ways: Firstly by breaking the `bridging points' between parts of the same cluster (to visualise this think of a cluster that looks like a figure of eight), which would cause overestimation  
of the number of the clusters (i.e. that cluster would become two smaller ones). Secondly by ignoring some clusters altogether, which would underestimate the number of clusters.

\subsection{Multi-level Conformal Clustering}\label{method}
\label{section:MLCC}

The conformal clustering approach described above only gives a snapshot of clustering tendency at one significance level. The problem with this is clear: at one level some data points may be classified as being within a virtual cluster, while at another level they may be classified as anomalies. A more thorough approach would involve performing CC at many levels and then evaluating the overall performance. \linebreak

\begin{algorithm}
\caption{Conformal Clustering at multiple levels}
\label{algo}
\begin{algorithmic}[1]
\STATE INPUT: $d$, dimension of the feature vector space $Z=R^d$
\STATE INPUT: data set $\{z_1,\dots,z_l\}\subset Z$
\STATE INPUT: NCM function $A: Z^{(*)}\times Z\rightarrow R$
\STATE SET a grid $G=\{g_1,\dots,g_u\}\in Z$ 
\STATE SET a grid $E=\{\varepsilon_1,\varepsilon_2,\dots,\varepsilon_{w-1},\varepsilon_w\}\subset[0,1]$
\FOR{ $i\in 1,2,\dots,u$ }
\STATE $z_{l+1}:=g_i$
\FOR{ $j\in 1,2,\dots,l+1$ }
\STATE $\alpha_j:=A\left(\{z_1,\dots,z_{i-1};z_{i+1},\dots,z_{l+1}\},z_i\right)$
\ENDFOR
\STATE $p_j:=\frac{|\{j\in 1,2,\dots,l+1:\alpha_j\ge\alpha_{l+1}\}|}{l+1}$
\ENDFOR
\FOR{ $i\in 1,\dots,w$ }
\STATE $e:=\varepsilon_i$
\STATE $R_e:=\{g_i\in G: p_i\ge e\}$
\STATE represent $R_e$ as a union of connected components $C_1(e),\dots,C_{k(e)}(e)$
\ENDFOR
\end{algorithmic}
\end{algorithm}

In this spirit, an implementation of a multi-level version of CC is shown in Algorithm~\ref{algo}. By $R^d$ here we mean the $d$-dimensional 
feature vector space.

The CP framework gives MLCC an exact coverage level, and so a stronger statistical meaning than standard clustering method. This means that a hypothethical new data point produced \textit{i.i.d.} from the same stochastic process as the observed data set will fall inside one of the virtual clusters identified by MLCC with probability $1-\varepsilon$. Thus a multi-level conformal cluster can be understood as a cluster with a flexible boundary: at the lowest significance level it is at its largest, while at the highest level it is at its tightest and has been reduced to its most `typical' examples, if it even exists at all by this point. \linebreak

We also note that the calculations at each point of the grid are independent of each other, meaning that they could easily be computed in a parallel fashion for improved scalability of the algorithm.

\section{Material and Methods}

\subsection{Real data sets and pre-processing}

The two datasets chosen for study are listed in Tab.~\ref{data_tab}. Both are taken from open UCI Machine Learning Repository.  \linebreak

The number of examples in the datasets have been reduced to 599 by random selection for improved plot visibility. Hence at each iteration of the loop through the grid points, 600 examples are compared with each other by the NCM. \linebreak

Given the computational burden of CC, which scales exponentially with the number of dimensions of the dataset, we are required to use some kind of dimensionality reduction method to render the computation of CC feasable in high dimensional cases. An example of this approach will be shown in the application of the MLCC technique on the HTRU dataset. Following~\cite{Cherubin2014}, we reduced it to two dimensions using the t-distributed Stochastic Neighbor Embedding (t-SNE) algorithm. This was done using the~\textbf{tsne} package in R (~\cite{tsne1} and~\cite{tsne2}). Reduction to two dimensions allows for improved visualization of the proposed clustering solution. \linebreak

The main advantage of t-SNE with respect to other non-linear dimensionality reduction techniques is that t-SNE takes, as the only input, the distance matrix between the points of a given dataset. This kind of approach ensures wide applicability of this method, including to non-standard data objects.
The main idea behind t-SNE is the reproduction of the connectivity structure of the original data: points that are ``similar" (close together according to the selected metric) in the original space  remain close together in the projection as well. Additional advantages of t-SNE over other methods such as Self-Organising Maps (SOMs) are also discussed in~\cite{tsne3}. \linebreak

Initially, each dataset also included a binary label for classification. We remind the reader that we do not use the label at the clustering stage, since clustering is an unsupervised technique. However, we do use it for evaluation of the clustering (after CC has been performed).  
\linebreak

\begin{table}
\begin{tabular}{|c|c|c|c|c|c|c|}\hline
No. & Name & Ex. & Feat.  & t-SNE & Grid & Labels \\\hline
1 & Skin  & 599 & 3 & No & 20x20x20 & Skin / Not Skin\\
2 & HTRU & 599 & 8 & Yes & 50x50 & Real Pulsar / Spurious \\
\hline
\end{tabular}
\caption{Data sets}\label{data_tab}
\end{table}

Features are rescaled depending on the number of dimensions in the dataset. For each of the features $x^j$ we apply the transformation
$$x^j\rightarrow S\frac{x^j-\min_{i}x_i^j}{\max_{i}x_i^j-\min_{i}x_i^j}$$
where $x_i^j$ is the value of $j$-th feature for $i$-th example,
$S=50$ if the overall number of features (dimensionality of the feature space) is 2, $S=20$ for dimensionality 3.

\subsection{Artificial noisy data sets}\label{artif1}

In addition to real data sets, we will perform some experiments on two-dimensional artificial data sets with varying levels of noise to test the performance and the robustness of the proposed method. \linebreak

Each of the data sets is a mixture sample (100 examples each) of five distributions:
\begin{enumerate}
\item a normal distribution with independent normal variables of same variance;
\item a normal distribution with independent normal variables of different variance;
\item a skewed normal distribution with dependent normal variables;
\item a circle distribution: the direction is random, the radius is distributed normally;
\item a half-circle arc distribution.
\end{enumerate}
Under existing degrees of freedom, the parameters of these distributions are chosen randomly.
In addition, some proportion of each sample  is corrupted in order to produce noise. The corruption
means that the variance is increased 5 times compared to the normal examples. \linebreak

\begin{figure}[H]	
\includegraphics[width=\linewidth]{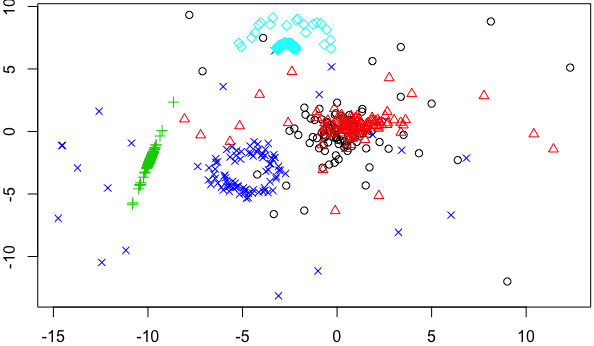}
\caption{Artificial random sample.
Each of the data sets is a mixture sample (100 examples each) of five distributions:
(black) a normal distribution with independent normal variables of same variance;
(red) a normal distribution with independent normal variables of different variance;
(green) a skewed normal distribution with dependent normal variables;
(blue) a circle distribution: the direction is random, the radius is distributed normally;
(sky) a half-circle arc distribution.
Noise level = 1/5: this is the proportion of artificial anomalies generated by same distributions as the clusters but with the variance increased 5 times.
Random seed=1.}\label{noise_05}
\end{figure}

Fig.~\ref{noise_05} shows one of the generated samples. 
In general, we will generate 15 of them:
five random parameter settings (seeds=1, 2, 3, 4, 5) and three versions of the noise proportion (1/10, 1/5 or 1/3).

\subsection{Conformal Clustering details} \label{CC-formal}

The underlying method is the $k$-nearest neighbour algorithm with the following NCM: \linebreak

$\alpha_i$ = Sum of the distances to the $k$ nearest neighbours. \linebreak

Once the region of conformity has been identified as explained in section \ref{method} it is divided into virtual clusters on the grid. Initially, we consider each grid point in the region as being its own individual virtual cluster. Then we merge clusters using a neighbouring rule, whereby two points $z_i$ and $z_j$ are in the same virtual cluster if they are within one grid point of each other on the grid. \linebreak

Finally, having established the virtual clusters, we project the observed dataset onto the grid and assign all points to various data clusters corresponding to the virtual cluster that they lie within. Any data points that do not lie in a virtual cluster are classified as anomalies.

\section{Results and Discussion}

\subsection{Multi-level conformal clustering results}

\begin{figure}[H]

{\bf Conformal clustering at $\varepsilon = 0.05$ (LHS) and $\varepsilon=0.2$ (RHS):}

{\bf 1) Skin dataset (all three dimensions against each other):}
		
\includegraphics[width=0.45\linewidth]{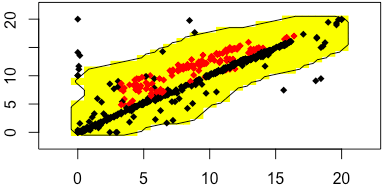}
\includegraphics[width=0.45\linewidth]{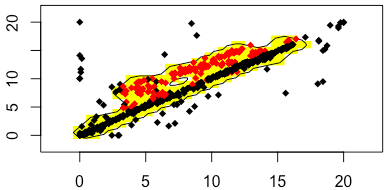}

\includegraphics[width=0.45\linewidth]{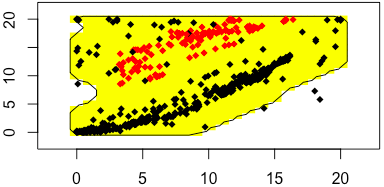}
\includegraphics[width=0.45\linewidth]{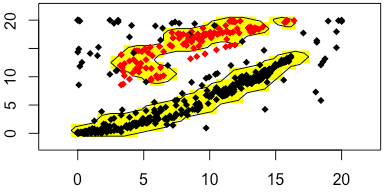}

\includegraphics[width=0.45\linewidth]{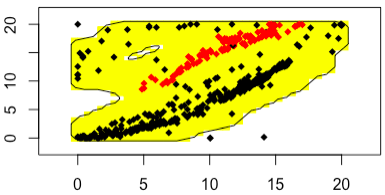}
\includegraphics[width=0.45\linewidth]{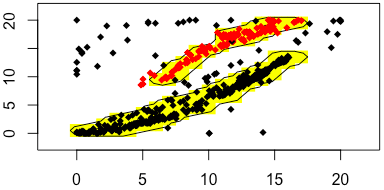}

{\bf 2) HTRU dataset (two t-SNE dimensions)}

\includegraphics[width=0.45\linewidth]{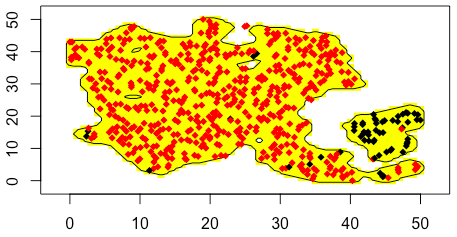}
\includegraphics[width=0.45\linewidth]{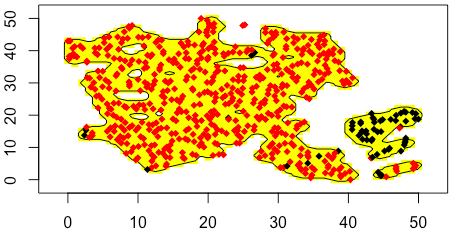}

		\caption{Scatter plots of both datasets featuring regions of conformity in yellow and labels in red/black. Left-hand column is at $\varepsilon = 0.05$ and right-hand column is at $\varepsilon = 0.2$.
		1) Skin: $x$-axis vs $y$-axis is blue feature vs green feature (top plots), blue vs red (middle plots), green vs red (bottom plots).
		2) HTRU: $x$-axis vs $y$-axis is t-SNE dimensions 1 vs 2 (no units).
		}
		
	\label{pred_sets}
	
\end{figure}

Fig.\ref{pred_sets} shows the application of CC to both real datasets, at $\varepsilon = 0.05$ on the left hand side and  $\varepsilon = 0.2$ on the right hand side. Dataset 1 had three dimensions so we have projected each dimension against the others to produce three plots for this dataset. Dataset 2 had eight dimensions which we reduced to two using the t-SNE algorithm, hence there is only one set of plots for this dataset. \linebreak

The data points themselves have been superimposed in red and black (corresponding to the binary labels) on the region of conformity which is shown in yellow. The connected components are the virtual clusters. The white region can be considered a `region of non-conformity'.
Table~\ref{words} explains the visual features of the plot. \linebreak

\begin{table}[H] \centering 
\begin{tabular}{|c|c|}\hline
{\bf Object} & {\bf Rpresented on the plot as} \\\hline
Features & Axes (two first features are shown)  \\\hline
Dataset & Red \& black points \\\hline
Region of conformity & Yellow region \\\hline
Region of non-conformity & White region  \\\hline
Virtual cluster & Individual yellow connected components  \\\hline
Data cluster & Red \& black points within a connected component \\\hline
Anomaly & Red or black points on white background  \\\hline
\end{tabular}
\caption{Explanation of the Conformal Clustering plots on Fig.~\ref{pred_sets}}\label{words}
\end{table}

Note how particularly for dataset 1 the coverage of the yellow region reduces as $\varepsilon$ increases from 0.05 to 0.2, which increases the number of anomalies identified. For dataset 2 this is less pronounced, and the lower value of $\varepsilon$ is actually enough for MLCC to identify the two classes in the data. For dataset 1 this is only possible at the higher value of $\varepsilon$.

\subsection{Visualising clustering over multiple significance levels in a dendrogram}\label{dendrogram}

\begin{figure}[H]

	\includegraphics[width=1\linewidth]{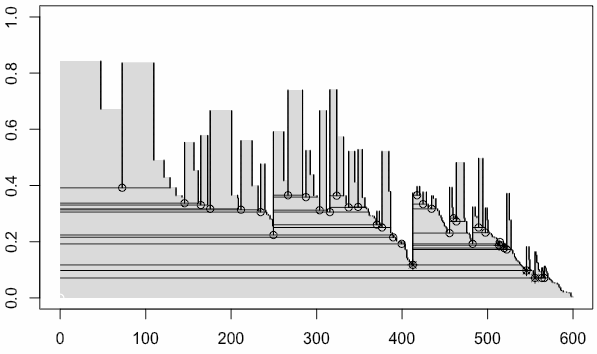}
	\includegraphics[width=1\linewidth]{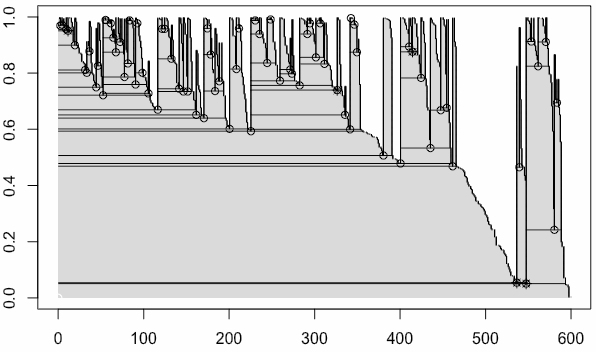}
	
	\caption{MLCC dendrograms for both datasets - dataset 1 (Skin) on top, dataset 2 (HTRU) on the bottom. $x$-axis represents data observation number (re-ordered according to cluster assignment), $y$-axis represents significance level}
	\label{dendro_plot}
	
\end{figure}

Fig.~\ref{dendro_plot} shows the dendrogram produced by MLCC. The $x$-axis represents all the observed
data examples in a convenient systematic order: examples belonging to the same virtual cluster at any significance level are placed next to each other. Sub-clusters of the same cluster are ordered by their size (left to right). Note that this makes the $x$-axis easier to interpret than in a dendrogram produced by HC, where the $x$-axis does not provide useful information about the data. For example, in dataset 2 the first split (which tells us about the most important partition of the data into clusters) is made between the examples 1--547, and 548--599. \linebreak

The $y$-axis represents significance level, and the colour tells us whether a data example lies within a virtual cluster (in which case it is coloured grey), or whether it is classified as an anomaly (white). Individual anomalies are left to the right of the cluster as the significance level increases (this is the best way to read the dendrogram - from bottom to top). \linebreak

For example, in the second dataset the first split happens at level $\varepsilon =  0.054$, producing three clusters. The largest of these (on the left-hand side) then reduces in size as points become anomalies until $\varepsilon =  0.469$ when the cluster splits up further. The clusters are represented as a kind of curvilinear trapezium: the bottom line of this trapezium for the largest cluster covers examples 1--547, while the upper line covers just examples 1--465 due to the loss of anomalies in  between.\linebreak

Sub-clusters of a divisible cluster are shown as standing on its `ceiling' - typically there are two of them as most splits are binary. The split points are marked on the figure as circles. Clusters which are indivisible at high values of $\varepsilon$ can be considered as containing the most `typical' examples of the observed data given the distribution thereof. \linebreak

\begin{figure}[H]

\includegraphics[width=0.8\linewidth]{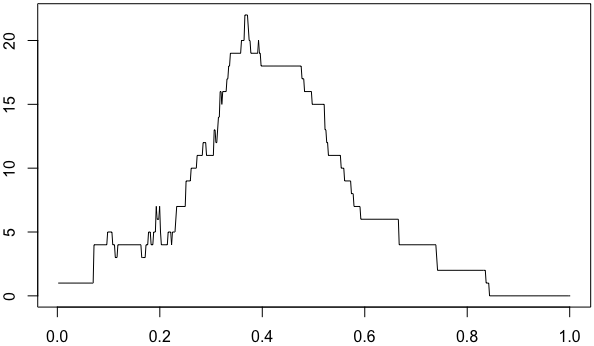}
\includegraphics[width=0.8\linewidth]{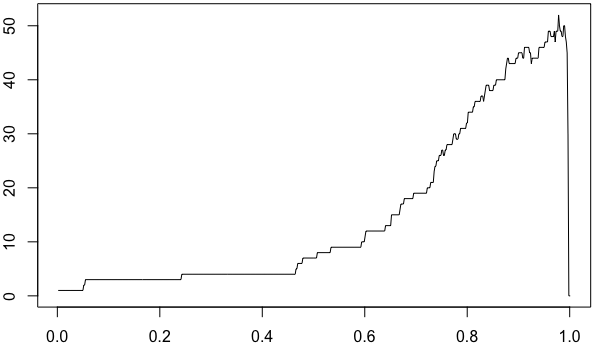}

		\caption{Number of clusters for different significance levels for both datasets - Skin on top, HTRU on bottom. $x$-axis represents significance level, $y$-axis represents number of clusters. Note how the curve has a non-monotonic shape, with a global maximum. This is the result of two opposing phenomena: an increase in $\varepsilon$ splits clusters until the less conforming points start to be classified as anomalies (and so they are ``removed" from the pool of clustered points). The position of the global maximum also yields information about the isotropy of the dataset: a maximum close to 0 means that the feature space has areas with a high density of data points, while a maximum close to 1 means that the density of the data points is more evenly spread out over the feature space so anomalies are identified only at higher significance levels.} 
	\label{num_clu}

\end{figure}

Fig.~\ref{num_clu} shows how the number of clusters depends on the significance level for each dataset. Typically it increases as $\varepsilon$ grows from zero because the large clusters split into smaller ones, but then drops back down again as $\varepsilon$ becomes so high that many clusters shrink to a size of zero and disappear altogether. See the caption for further explanation.

\subsection{Conceptual comparison of MLCC and HC}

We suggest that MLCC can compete with more traditional methods of clustering such as $k$-means and hierarchical clustering (HC). Since MLCC shares many features with HC, we consider this as the primary point of comparison \linebreak

Agglomerative HC is based on the merging together of the closest-together examples. Starting with a completely disjoint set, the clusters are gradually merged until the whole dataset is just one cluster. At each stage the algorithm finds the pair of existing clusters with the smallest distance between them (the distance between two clusters is defined as
the smallest distance between their constituents). \linebreak

This method can also be understood in the opposite way - as moving from one cluster to a completely disjointed picture, by selecting the most `high-distance split' on each step (omitting details here). This understanding lets us draw a parallel with increasing the significance level in MLCC. \linebreak

\begin{figure}[H]	

{\bf HC (LHS) and MLCC (RHS) applied to HTRU dataset}

	\includegraphics[width=0.5\linewidth]{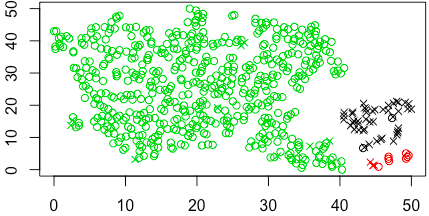}
	\includegraphics[width=0.5\linewidth]{cc05_20pc.png}
		
	\caption{The three clusters solution of HC vs. CC at $\varepsilon=0.2$, HTRU data.
	Left plot: HC, the colour is for the cluster number, the shape is for the label.
	Right plot: MLCC, the clusters are in yellow, the red/black colour is for the label.}
	\label{hc_hc}
\end{figure}

The HC solution with three clusters is shown on Fig.~\ref{hc_hc}, compared with the MLCC clustering for  $\epsilon=0.2$.
On the left plot, different shapes represent different labels, while the different colours identify the three different clusters. We can see that the shape and the content of the clusters are essentially the same for both methods, with the very notable exception of a group of outliers in MLCC in the mid-top part of the right plot. 

\subsection{Numerical comparison of MLCC and HC}

To evaluate MLCC, and compare it numerically with traditional HC, we use purity of clusters as the main criterion. This is an applicable metric when the true labels of the data are known. Here we take these labels from the original dataset and convert them to binary ones if needed, using the average (mean) values as a threshold. The purity of a cluster (that is, a data cluster, not a virtual one) means the largest proportion of its instances belonging to the same class. \linebreak

We calculate purity for the clusters that appear during the 10 first splits - i.e. the 20 largest clusters.
Tab.~\ref{table:hc_purity} compares the purity of CC and HC in the following way: In both cases, we have taken the 20 first clusters, and averaged their purity.
In both datasets MLCC shows better performance according to this criterion. 
\begin{table}[H]
	\centering
	\caption{Comparison of purity for multi-level CC vs HC}\label{table:hc_purity}
\begin{tabular}{|c|c|c|} \hline
averaged purity	& Multi-level CC & Traditional HC \\\hline
Skin                  &  0.965 & 0.925 \\\hline
HTRU               & 0.954  & 0.935 \\\hline   \end{tabular}
\end{table}

\begin{figure}[H]	
\caption{The results for: noise=1/5, seed=1. 
The plot includes the clustering for
$\varepsilon=0.2$, and the clustering tree
aggregating all the levels.}\label{noise_05a}
\includegraphics[width=\linewidth]{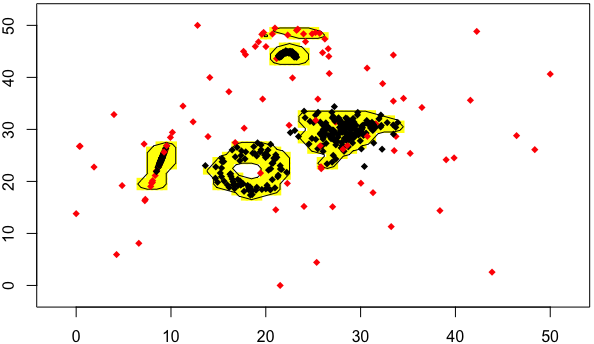}

\includegraphics[width=\linewidth]{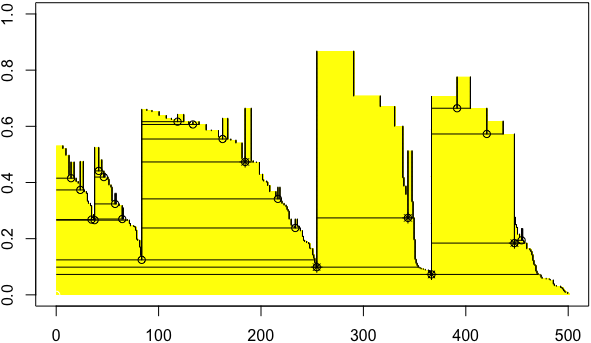}
\end{figure}

\subsection{Clustering and Anomaly Detection on noisy artificial data}\label{artif2}

Experiments in this section are performed on the artificial noisy data examples described in Sec.~\ref{artif1}. A sample result is shown in Fig.~\ref{noise_05a}. It corresponds to the data in Fig.~\ref{noise_05}.
The red points are the noisy examples generated with increased variance, the black points are the normal examples. \linebreak

The yellow area is the region of conformity for $\varepsilon=0.2$. For perfect anomaly detection we would expect this region to contain all the black points and exclude all the red points. As we can see, there are a few mistakes for both classes but the overall performance is good. By modifying the significance level we can calculate the AUC (area under curve) value, which is 0.89 for this particular example. We remind the reader that an AUC value close to 1 indicates very good classifier. 
The Table~\ref{auc} presents the summary of AUC achieved in the different settings.
The full clustering pictures for all combinations of noise levels and seeds, are given in Appendix.

\begin{table}[H]
	\centering
	\caption{Clustering and Anomaly Detection: noise level 1/10, seeds 1--5}\label{auc}
\begin{tabular}{|c|ccc|} \hline
AUC & noise 1/10 & noise 1/5 & noise 1/3 \\\hline
seed=1 & 0.87 & 0.89 & 0.82\\
seed=2 & 0.89 & 0.89 & 0.90\\
seed=3 & 0.72 & 0.74 & 0.60\\
seed=4 & 0.84 & 0.84 & 0.75\\
seed=5 & 0.81 & 0.65 & 0.65\\\hline
\end{tabular}
\end{table}

Summarising the results, we see that the averaged (by seeds) AUC is 0.83 for noise level 1/10, 0.80 for 1/5 and 0.74 for 1/3. We remind the reader that by noise level we refer to the proportion of anomalies imputed into the data set. Naturally the higher this is the worse the performance of the method, but we note that there is not a significant drop in AUC even when the noise level is more than tripled from 1/10 to 1/3. \linebreak

The results also show that the clusters identified in the region of conformity mostly match either the original clusters or their overlapping unions. An exception to this is the fifth (half-circle arc) distribution: In some of the figures its noisy background is identified as a separate cluster, or as a non-anomalous part of the same cluster. This is the consequence of imputing an anomalous class that in other circumstances may be simply another normal class. One of the advantages of combining anomaly detection with clustering is the attention given to such data points: further analysis of a small cluster may lead to conclusion of its abnormality.

\section{Conclusion}

In this study we have developed, described and tested the Multi-Level Conformal Clustering (MLCC) methodology, a variation on the Conformal Clustering algorithm that introduces multiple significance levels. We propose MLCC as a viable alternative to hierarchical clustering with unique advantages of its own. In particular MLCC combines clustering and anomaly detection in a CP framework that is able to provide statistical robustness. \linebreak

We have applied the technique to both real world and artificial datasets, and shown that MLCC is able to produce dendrograms which are both more informative and easier to interpret than those produced by HC. We have also compared the techniques numerically, with MLCC showing improved performance over HC. \linebreak

There remain some interesting questions to be analysed within the realm of Conformal Clustering. The first involves running the techique with more underlying clustering algorithms, such as self-organizing maps or support-vector machine-based clustering. \linebreak

Another key point is performance: the speed of MLCC is very dependent on the size of the grid representing the feature space, which grows exponentially with the number of features in the data to be clustered. The solutions to this problem are along two lines of research: the first being related to an increase in the speed of the algorithm, for instance exploiting the embarassingly parallel nature of the evaluation step. The second one is related to using a data-driven grid, instead of a regular lattice, for algorithm evaluations. \linebreak

One more line of research involves scenarios when there are overlapping clusters. At one significance level we may find that two clusters overlap, while at another they may have got smaller and are now disjoint. This tendency is worthy of further investigation, as it differs from the way that clusters are split in, for example, hierarchical clustering.

\section*{Acknowledgements}

This research was funded by Centrica plc company and Royal Holloway University of London.
Additionally, it was funded by AstraZeneca and Mittie companies.


\section*{Appendix. Results on Artificial Data}

\begin{figure}[H]
  \includegraphics[width=0.8\linewidth]{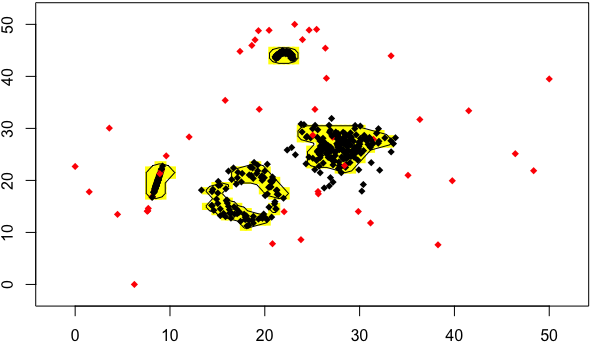} 
  
  \includegraphics[width=0.8\linewidth]{b051.png} 
  
  \includegraphics[width=0.8\linewidth]{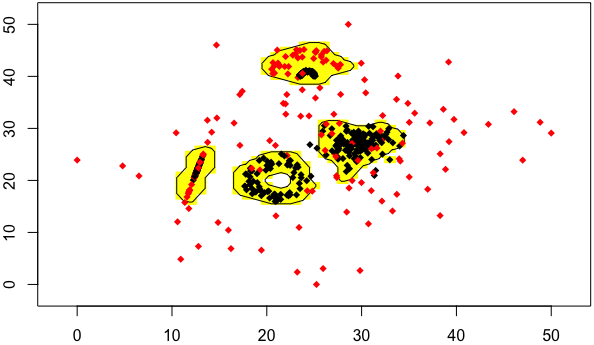} 
\caption{Clustering plots for seed=1, noise levels = 1/10, 1/5, 1/3}
\label{app1}  
\end{figure}

\begin{figure}[H]
  \includegraphics[width=0.8\linewidth]{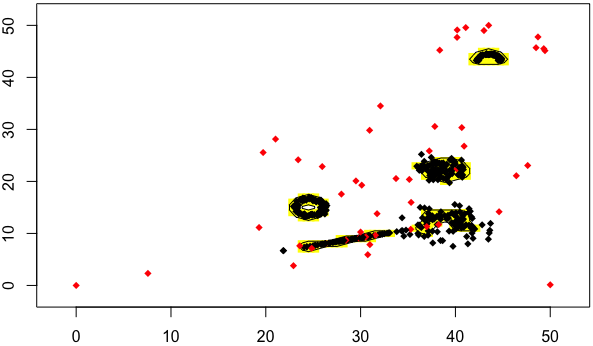} 
  
  \includegraphics[width=0.8\linewidth]{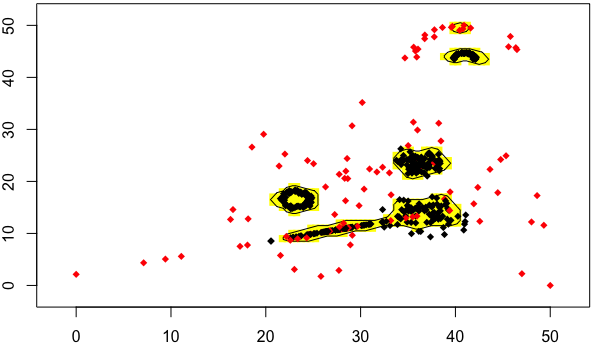} 
  
  \includegraphics[width=0.8\linewidth]{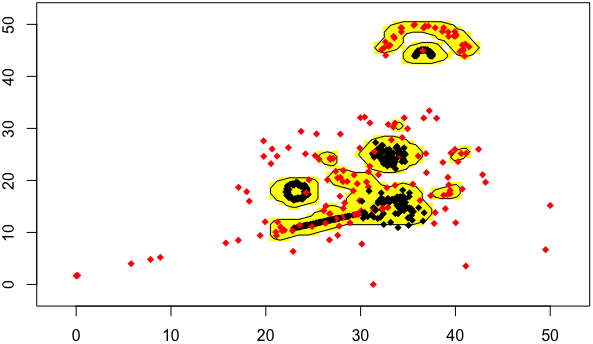} 
\caption{Clustering plots for seed=2, noise levels = 1/10, 1/5, 1/3}
\label{app2}  
\end{figure}

\begin{figure}[H]
  \includegraphics[width=0.8\linewidth]{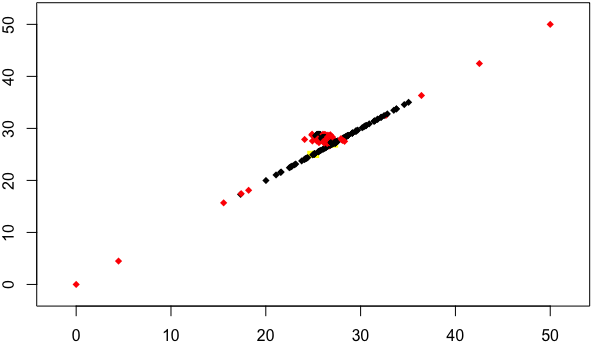} 
  
  \includegraphics[width=0.8\linewidth]{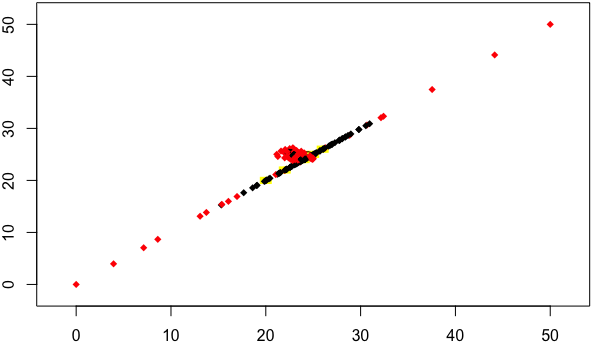} 
  
  \includegraphics[width=0.8\linewidth]{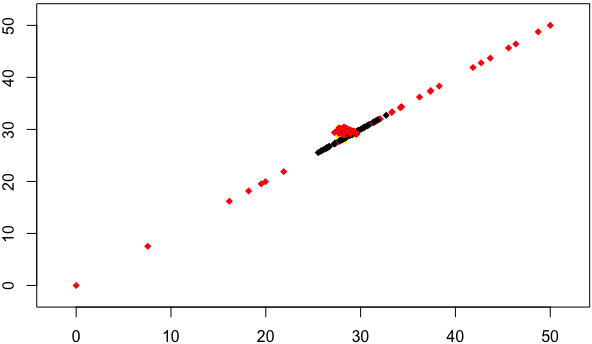} 
\caption{Clustering plots for seed=3, noise levels = 1/10, 1/5, 1/3}
\label{app3}  
\end{figure}

\begin{figure}[H]
  \includegraphics[width=0.8\linewidth]{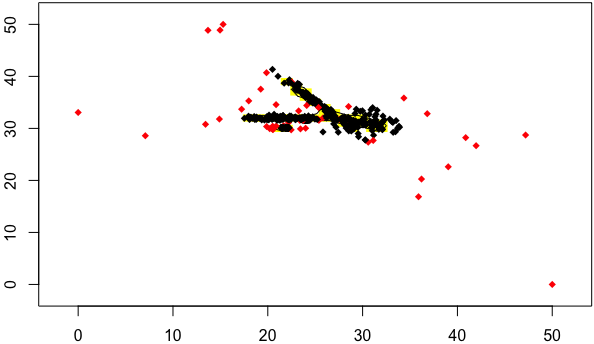} 
  
  \includegraphics[width=0.8\linewidth]{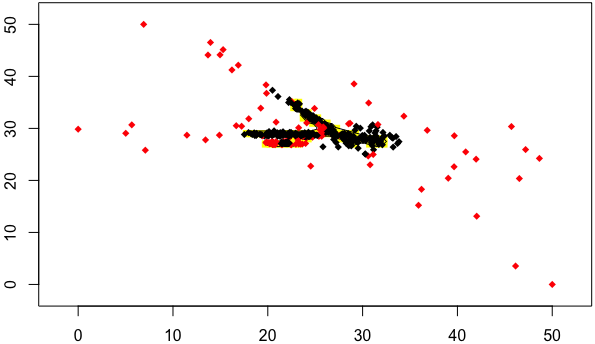} 
  
  \includegraphics[width=0.8\linewidth]{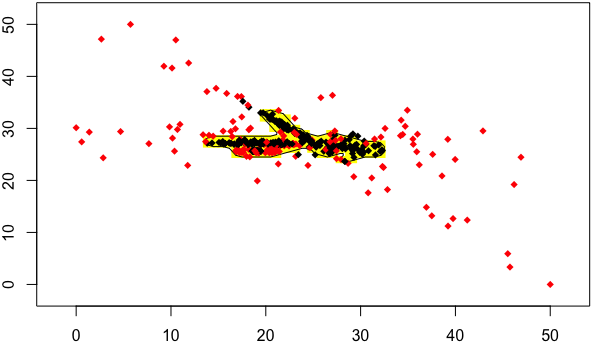} 
\caption{Clustering plots for seed=4, noise levels = 1/10, 1/5, 1/3}
\label{app4}  
\end{figure}

\begin{figure}[H]
  \includegraphics[width=0.8\linewidth]{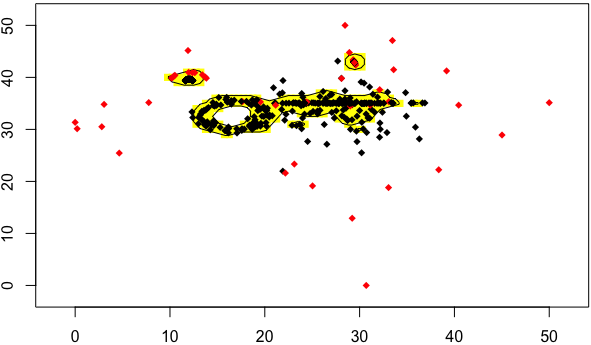} 
  
  \includegraphics[width=0.8\linewidth]{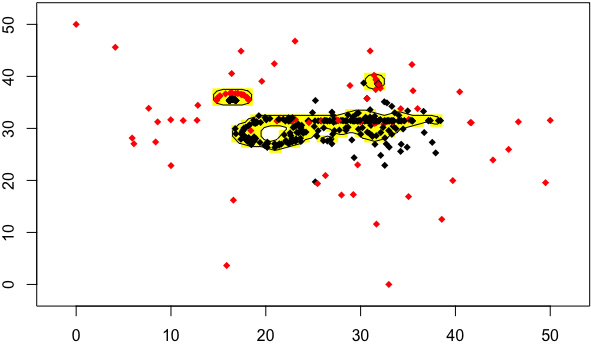} 
  
  \includegraphics[width=0.8\linewidth]{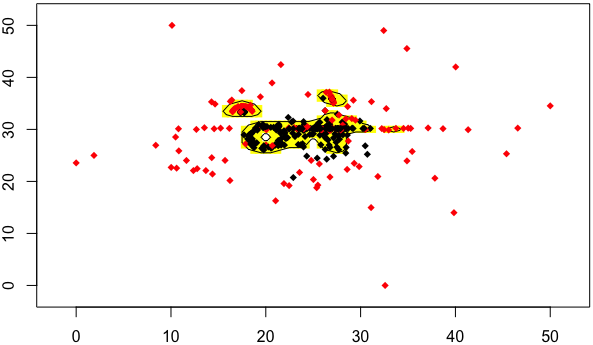} 
\caption{Clustering plots for seed=5, noise levels = 1/10, 1/5, 1/3}
\label{app5}  
\end{figure}

Here we present the results for the versions of the artificial data, created for different random seeds and noise levels,
as described in Sec.~\ref{artif1}.
Figures~\ref{app1}--\ref{app5} correspond to the seeds 1--5 respectively,
and each of them presents three noise level versions.

\end{document}